\pdfoutput=1

\documentclass[11pt]{article}

\newcommand{\repourl}{https://github.com/GGLAB-KU/turkish-plu}

\usepackage[]{acl}

\usepackage{times}
\usepackage{latexsym}
\usepackage{xcolor}
\usepackage{amssymb}

\usepackage[T1]{fontenc}
\usepackage{multirow}
\usepackage{booktabs}
\usepackage{pgfplots}
\pgfplotsset{compat=1.18}

\usepackage[utf8]{inputenc}

\usepackage{microtype}
\usepackage{inconsolata}
\usepackage{amsmath}
\usepackage{graphicx}
\usepackage{tabularx}
\usepackage{makecell}

\usepackage{todonotes}

\graphicspath{ {./Images/} }
\title{Benchmarking Procedural Language Understanding for Low-Resource Languages: A Case Study on Turkish}

\author{Arda Uzunoğlu\textsuperscript{$\ddagger$}\thanks{The work was done while the first author was at Eskişehir Bahçeşehir College.} ,~
Gözde Gül Şahin\textsuperscript{$\dagger$}~
\\[.3em]
\textsuperscript{$\ddagger$}Computer Science Department, Johns Hopkins University, Maryland, USA \\
\textsuperscript{$\dagger$}Computer Engineering Department, Koç University, Istanbul, Türkiye\\
\textsuperscript{$\dagger$}{\url{https://gglab-ku.github.io/}}\\
}

\begin{document}
\maketitle
\begin{abstract}
Understanding procedural natural language (e.g., step-by-step instructions) is a crucial step to execution and planning. However, while there are ample corpora and downstream tasks available in English, the field lacks such resources for most languages. To address this gap, we conduct a case study on Turkish procedural texts. We first expand the number of tutorials in Turkish wikiHow from 2,000 to 52,000 using automated translation tools, where the translation quality and loyalty to the original meaning are validated by a team of experts on a random set. Then, we generate several downstream tasks on the corpus, such as linking actions, goal inference, and summarization. To tackle these tasks, we implement strong baseline models via fine-tuning large language-specific models such as TR-BART and BERTurk, as well as multilingual models such as mBART, mT5, and XLM. We find that language-specific models consistently outperform their multilingual models by a significant margin across most procedural language understanding~(PLU) tasks. We release our corpus, downstream tasks and the baseline models with \url{\repourl}.
\end{abstract}


\section{Introduction}

A procedural text typically comprises a sequence of steps that need to be followed in a specific order to accomplish a goal. For example, to care for an indoor plant, one must undertake tasks such as i) \textit{selecting an appropriate location for the plant}, ii) \textit{maintaining indoor humidity levels}, and iii) \textit{selecting the right fertilizer}, usually in the given order. To accomplish a goal given with step-by-step instructions, a set of diverse skills that can be related to traditional NLP tasks such as semantic analysis (e.g., who did what to whom), commonsense reasoning (e.g., plant requires water), and coreference resolution (e.g., \textit{it} refers to the \textit{plant}) are required. Hence, procedural language understanding~(PLU) can be considered a proxy to measure the performance of models on a combination of these distinct skills. 


Previous work has immensely utilized the WikiHow tutorials, and proposed several downstream tasks on procedural text. For example, \citet{zhang-etal-2020-reasoning} introduced step and goal inference tasks where the objective is to predict the most likely \textit{step} given the \textit{goal} or vice versa. Similarly, \citet{zellers-etal-2019-hellaswag} proposed predicting the \textit{next event} given the goal and the current step. All of these tasks are formulated as multiple-choice QA and require a partial understanding of step-goal relations in procedural documents. Furthermore, \citet{zhou-etal-2022-show} proposed an information retrieval task where the goal is to link \textit{steps} to related \textit{goals} to create a wikiHow hierarchy. Finally, several other works~\citep{Koupaee2018WikiHowAL, ladhak-etal-2020-wikilingua} proposed an abstractive summarization task, that requires competitive language generation skills.  


Despite its importance, PLU has been largely ignored for the majority of the languages due to a lack of language-specific web corpora. Except from \citet{ladhak-etal-2020-wikilingua}, all the aforementioned tasks are only available in English. In addition to the scarcity of raw text, creating downstream task data is challenging and might require language-specific filtering techniques to ensure high quality. Finally, all previous works study the proposed tasks in isolation, which can only give a limited insight into the model's performance. 


Considering the uneven distribution of available procedural data across languages\footnote{Although wikiHow comprises 19 languages, only two languages (English and Spanish) have more than 100k articles in parallel \citep{ladhak-etal-2020-wikilingua}.}, our objective is to inspire research efforts on PLU for other understudied languages from different language families. To achieve this, we design a case study focused on the Turkish language. Unlike previous works, we adopt a centralized approach and introduce a comprehensive benchmark that contains six downstream tasks on procedural documents. 

To address the scarcity of resources, we utilize automatic machine translation tools. We implement rigorous quality control measures for machine translation including human evaluation, and show that the data is indeed high-quality. Next, we survey and study several downstream tasks and create high-quality, challenging task data through language-specific filtering and manual test data annotation. Finally, we perform a comprehensive set of experiments on a diverse set of language models with different pretraining, fine-tuning settings, and architectures. We find that language-specific models mostly outperform their multilingual counterparts; however, the model size is a more important factor than training language, i.e., large enough multilingual models outperform medium sized language-specific models. We show that tasks where we can perform rigorous language-specific preprocessing such as goal inference, are of higher-quality, thus more challenging. Finally, we find that our best-performing models for most downstream tasks, especially reranking, goal inference, and step ordering, are still far behind their English counterparts, suggesting a large room for improvement. We release all the resources---including the structured corpus of more than 52,000 tutorials, data splits for six downstream tasks and the experimented baseline models--- at \url{\repourl}.

\section{Related Work}
    WikiHow is an eminent source for studying procedural text, allowing for a broad range of NLP tasks to be proposed and studied, such as linking actions \citep{lin-etal-2020-recipe, zhou-etal-2022-show}, step and goal inference \citep{zhang-etal-2020-reasoning, yang-etal-2021-visual}, step ordering \citep{zhang-etal-2020-reasoning, zhou-etal-2019-learning-household}, next event prediction \citep{nguyen-etal-2017-sequence, zellers-etal-2019-hellaswag}, and summarization \citep{Koupaee2018WikiHowAL, ladhak-etal-2020-wikilingua}. While these works serve as a proxy to procedural text understanding, they are mostly limited to English.
    
    Exploiting machine translation tools is a common practice to generate semantic benchmarks for many resource-scarce languages. For instance, \citet{mehdad-etal-2010-towards} automatically translated hypotheses from English to French to generate a textual entailment dataset. Similarly, \citet{10.1007/978-3-319-99722-3_31} created a Portuguese corpus for natural language inference (NLI), namely as SICK-BR, and \citet{unknown} introduced the first Swedish benchmark for semantic similarity, by solely employing automatic translation systems. 
    Moreover, \citet{budur-etal-2020-data} and \citet{beken-fikri-etal-2021-semantic} employed Amazon and Google translate to generate Turkish NLI and sentence similarity, datasets via automatically translating existing resources such as SNLI \citep{bowman-etal-2015-large}, MNLI \citep{williams-etal-2018-broad} and STS-B \citep{cer-etal-2017-semeval}.

    
\section{Turkish PLU Benchmark}
\label{section:tasks}

    To evaluate the procedural language understanding capacity of existing models and to improve upon them, we introduce i) a large procedural documents corpus covering a wide range of domains for Turkish, ii) a diverse set of downstream tasks derived from the corpus to evaluate distinct large language models and iii) strong baselines for each task.

    \subsection{Corpus}
        Following previous work~\citep{zhang-etal-2020-reasoning}, we utilize wikiHow, a large-scale source for procedural texts that contains how-to tutorials in a wide range of domains, curated by experts. 
        We follow the format used by \citet{zhang-etal-2020-reasoning} and extract the title, methods/parts, steps, and additional information, such as the related tutorials, references, tips, and warnings. 
        We focus on the categories with the least subjective instructions (e.g., Crafts) and ignore subjective categories (e.g., Relationships). 
        
        Our corpus creation process has two steps: i) scraping the original Turkish wikiHow, and ii) translating the English tutorials from the English wikiHow corpus~\citep{zhang-etal-2020-reasoning}. 
        
        \paragraph{Scraping Turkish Wikihow} Using the beautifulsoup library \citep{richardson2007beautiful}, we scrape the Turkish wikiHow tutorials from the sitemap files. After the category filtering and deduplication process, we get over 2,000 tutorials.
        
        
        \begin{table}[!htp]
        \centering
        \scalebox{0.7}{
        \begin{tabular}{rrrrrr}
            \textbf{BLEU} & \textbf{ROUGE} & \textbf{METEOR} & \textbf{COMET} & \textbf{chrF} & \textbf{chrF++} \\
            \midrule
            23.51 & 52.25 & 44.32 & 88.12 & 67.91 & 62.08   
        \end{tabular}
        }
        \caption{BLEU, ROUGE, METEOR, COMET, chrF, and chrF++ scores calculated over 1734 translated English-Turkish article pairs. All of the metrics are mapped to the interval of [0, 100] for convenience. Higher score indicates better translation for each evaluation metric.}
        \label{table:mtqt}
        \end{table}

        \begin{table}[!htp]
        \centering
        \scalebox{0.7}{
        \begin{tabular}{rrrrrrr}
            \textbf{} & \textbf{Fleiss' Kappa} & \textbf{Average} & \textbf{Agree 5} & \textbf{Agree +4} \\
            \midrule
            i) & 0.751 & 4.40 & 47\% & 69\% \\
            ii) & 0.813 & 4.76 & 78\% & 87\% \\
        \end{tabular}
        }
        \caption{Results of the expert human validation on automatic machine translation quality control. Agree 5 and +4 respectively represent the percentage of the experts who agree that the score must be 5 or 4 and more.}
        \label{table:mtqhv}
        \end{table}
    
        \paragraph{Translating the English Wikihow} 
        To automatize the translation process, we first develop an open-source \textit{file-level} translation tool: \textsc{ÇeVeri}. 
        It is simply an easy-to-use Google Translate\footnote{\href{https://cloud.google.com/translate}{https://cloud.google.com/translate}} wrapper that utilizes recursive search to find, translate and replace nested text fields within a file~(see Appendix~\ref{sec:ceveri}). After filtering the subjective categories, we translate over 50,000 tutorials using \textsc{ÇeVeri}. 
        
        \paragraph{{MT Quality Control}} To measure the translation quality of \textsc{ÇeVeri}, we translate the English counterparts of the original Turkish wikiHow tutorials and calculate a set of automatic evaluation metrics such as BLEU and COMET~\citep{papineni-etal-2002-bleu, lin-2004-rouge, banerjee-lavie-2005-meteor, rei-etal-2020-comet, popovic-2015-chrf} given in Table~\ref{table:mtqt}. Although we use conventional metrics such as BLEU to align well with the literature, we are aware of the concerns related to them \citep{freitag-etal-2022-results}. Therefore, we include metrics that better correlate with human evaluations, such as COMET~\citep{mathur-etal-2020-results, freitag-etal-2021-results}, and consider character-level information such as chrF~\citep{popovic-2015-chrf}. Considering these, \textsc{ÇeVeri} achieving considerably high COMET and chrF scores indicate that the translation is, indeed, of high quality.


        We also conduct human validation with three native Turkish speakers fluent in English. We randomly sample 104 step triplets: a) the original Turkish step, b) the corresponding English step, and c) the translation of the English step with respect to the category distribution of our corpus. Each expert is asked to evaluate the triplets by i) scoring the translation quality with the English step and the translated Turkish step and ii) scoring the semantic similarity between the original and the translated Turkish steps both between 1 and 5 (inclusive; 5 is the best). As given in Table~\ref{table:mtqhv}, the results are highly reassuring, indicating high average scores with substantial agreement \citep{Fleiss1971MeasuringNS}. Additionally, we perform a pilot study to investigate the feasibility of using machine-translated data and find that silver data bring a noticeable improvement~(see Appendix \ref{sec:feasibility_mt}). Therefore, we consider the automatically generated part of our corpus to be of high quality due to the results of both the automatic and manual quality controls and the pilot study.
        
        
        
        \paragraph{Corpus Statistics} 
        \begin{table}
            \begin{center}
            \scalebox{0.65}{
            \begin{tabular}{lrrr}
            \toprule
            \multirow{2}{*}{\textbf{Source}} & \multirow{2}{*}{\textbf{\#Tutorials}} &\textbf{\#Steps} & \textbf{\#Methods}\\
            & & \textbf{Avg Steps} & \textbf{Avg Methods} \\ 
            \midrule
            \multirow{2}{*}{C\&OV} & \multirow{2}{*}{2K} & 32K & 5K \\
            & & 13.42 & 2.33\\
            \multirow{2}{*}{C\&E} & \multirow{2}{*}{16K} & 229K & 34K \\
            & &13.89 & 2.10\\
            \multirow{2}{*}{HE} & \multirow{2}{*}{11K} & 154K & 31K \\
            & &14.34 & 2.87 \\
            \multirow{2}{*}{H\&C} & \multirow{2}{*}{9K} & 119K & 19K \\
            & &13.37 & 2.20 \\
            \multirow{2}{*}{H\&G} & \multirow{2}{*}{10K} & 133K & 25K \\
            & &13.66 & 2.59 \\
            \multirow{2}{*}{P\&A} & \multirow{2}{*}{4K} & 53K & 11K \\
            & &13.75 & 2.86 \\
            \midrule
            \multirow{2}{*}{Original} & \multirow{2}{*}{2K} & 38K & 7K \\
            & & 19.15 & 3.35 \\
            \multirow{2}{*}{Translated} & \multirow{2}{*}{50K} & 681K & 120K \\
            & & 13.61 & 2.40 \\
            \bottomrule
            \multirow{2}{*}{\textbf{Total}} & \multirow{2}{*}{\textbf{52K}} & \textbf{719K} & \textbf{127K} \\
            & & \textbf{13.83} & \textbf{2.43} \\
            \bottomrule
            \end{tabular}
            }
            \caption{Final corpus statistics. C\&OV: Cars and Other Vehicles, C\&E: Computers and Electronics, HE: Health, H\&C: Hobbies and Crafts, H\&G: Home and Garden, P\&A: Pets and Animals. Avg Step and Method: Average number of steps and methods per tutorial, respectively. A method is a set of steps that can be followed to achieve the given goal, while a step is a single instruction.}
            \label{table:corpus_stats}
            \end{center}
        \end{table}
        Our final corpus has more than 52,000 tutorials from six wikiHow categories, which contain around 719K steps and around 127K methods, with an average of 13.83 steps and 2.43 methods per tutorial as given in Table~\ref{table:corpus_stats}. Computers and Electronics is the largest category, while the Cars and Other Vehicles is the smallest. We posit the number of tutorials for a category decreases as the level of expertise needed for writing tutorials for that category increases. Health category is an exception to this, as most of its articles do not really go into depth, and contain basic and simple instructions. Although average numbers of steps and methods per tutorial are consistent by categories, they vary by data creation methods. We believe the reason for such a difference is that the tutorials translated and added to Turkish wikiHow by editors are far more popular and gripping tutorials, which probably correlates with the level of ease, thus the descriptiveness and comprehensiveness, of the tutorials. We hypothesize that they are prioritized in the translation line by wikiHow editors, as they attract more attention.

    \subsection{Downstream Tasks}
    \label{subsection:downstream}
    
        Next, we inspire from previous works that studied a single downstream task created on wikiHow and combine them under a single benchmark, summarized in Table \ref{table:task_specific_datasets} and explained below.  
        \begin{table}
            \begin{center}
            \scalebox{0.85}{
            \begin{tabular}{lrrr}
            \toprule
            \textbf{Task} & \textbf{Train} &\textbf{Validation} & \textbf{Test}\\
            \midrule
            Linking Actions & 1319 & \textemdash & 440 \\
            Goal Inference & 255K & 5K & 837 \\
            Step Inference & 124K & 5K & 612 \\
            Step Ordering & 539K & 10K & 1021 \\
            Next Event Prediction & 82K & 5K & 656 \\
            Summarization & 113K & 6K & 6K \\
            \bottomrule
            \end{tabular}
            }
            \caption{Downstream tasks and dataset split sizes.}
            \label{table:task_specific_datasets}
            \end{center}
    \end{table}
    
    \paragraph{Linking Actions}
    
\begin{figure}[h]
            \centering
            \includegraphics[width=0.5\textwidth]{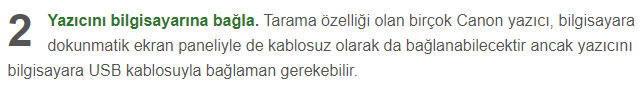}
            \caption{An example step with a hyperlink redirecting it to a tutorial. (Step says ``Connect your printer to your computer" and the redirected tutorial has the title of ``How to Connect a Printer to a Computer")}
            \label{figure:step_with_a_hyperlink}
            \end{figure}

    The task is defined as detecting the links between the steps and the goals across articles as shown in Figure \ref{figure:step_with_a_hyperlink}. 
    The steps provided in the tutorials, along with their hyperlinked goals, serve as the ground-truth data for the linking actions task.

    \paragraph{Goal Inference}
    The goal inference task is simply defined as predicting the most likely goal, given a step. This task is structured as a multiple-choice format~\citep{zhang-etal-2020-reasoning}. For instance, when the prompt step is "Kıyafetlerini sık, böylece daha hızlı kuruyacaktır. (Squeeze your clothes, they would get dry quicker this way.)" and the candidate goals are: \\
            A. Lavanta Nasıl Kurutulur? (How to Dry Lavender) \\
            B. Kıyafetler Elde Nasıl Yıkanır? (How to Hand-Wash Clothes) \\
            C. Kıyafetler Çabucak Nasıl Kurutulur? (How to Dry Clothes Quickly) \\
            D. Islak Bir iPhone Nasıl Kurutulur? (How to Dry a Wet iPhone) \\
            then the answer would be \textbf{C}. \\
    We collect the positive step-goal pairs by iteratively picking them from each tutorial. For the negative candidate sampling, we consider both the semantic similarity with the positive candidate and the contextual plausibility for the step. We first encode each step in our corpus by averaging the BERT embeddings~\citep{devlin-etal-2019-bert} of the verb, noun, and proper noun tokens~\footnote{We conduct the POS tagging with the nlpturk library. \href{https://github.com/nlpturk/nlpturk}{https://github.com/nlpturk/nlpturk}} contrary to \citet{zhang-etal-2020-reasoning}, which only considers the verb tokens. The reason why we include the additional POS tags is that most of the steps and goals in our corpus contain auxiliary verbs, which are common to Turkish such as \textit{``yemek yapmak''}~(to cook)\footnote{Such auxiliary verbs are mainly etmek, eylemek, olmak, kılmak and yapmak.}. Although contextualized embeddings help distinguish such differences to a certain extent, we observe that the incorporation of the additional parts brings a significant improvement in our negative candidate sampling strategy. Using FAISS~\citep{johnson2019billion} with the our vector representations, we choose the top-3 goals with the highest cosine similarity to the positive candidate as the negative candidates. After the positive and negative candidate sampling, we randomly reassign one of the candidates as positive and correct the labels accordingly with a probability of 0.15 to avoid the model learning the sampling strategy. Lastly, we apply a set of hand-crafted filters~\citep{zhang-etal-2020-reasoning} to ensure the quality of the task-specific data.
    
    \paragraph{Step Inference}
    Similar to the goal inference task, step inference is defined as predicting the most likely goal for the given step. It is also formulated as a multiple choice task~\citep{zhang-etal-2020-reasoning}. For instance, when the prompt goal is ``Makas Nasıl Bileylenir? (How to Whet a Scissors)'' and the candidate steps are: \\
            A. Camı temizle. (Clean the glass/windows.) \\
            B. Makası sil. (Wipe the scissors.) \\
            C. Tuvaleti sil. (Wipe the toilet.) \\
            D. Kartonu kes. (Cut the cardboard.) \\
            the answer would be \textbf{B}. \\
    We follow the same steps as in goal inference to sample positive and negative candidates by simply reversing the roles of the goals and the steps in the sampling process.
    
    \paragraph{Step Ordering} Here, the goal is to predict the preceding step out of the two given steps that help achieve a given goal. Similarly, it is formulated as a multiple-choice task. For instance, when the prompt goal is
            ``YouTube'da Nasıl Yorum Bırakılır? (How to Leave a Comment on Youtube)'' and the candidate steps are: \\
            A. Bir video arayın. (Search for a Video.) \\
            B. YouTube'u açın. (Open Youtube.) \\
            \textbf{B} would be the answer since it must precede A.\\
    For this task, we use the sampling strategy of \citep{zhang-etal-2020-reasoning}. In wikiHow, some tutorials follow an ordered set of steps, while others contain alternative steps parallel to each other. Out of the ordered portion of our corpus, obtained in Appendix \ref{sec:classifying_orderliness_of_the_tutorials}, we use each goal as a prompt to sample step pairs with a window size of 1 and do not include any non-consecutive steps. We also randomly shuffle the pairs to prevent any index biases.
    
    \paragraph{Next Event Prediction}
    This task aims to produce the following action for a given context. It can be formulated as either a text generation task~\citep{nguyen-etal-2017-sequence, zhang-etal-2020-analogous} or a multiple-choice task~\citep{zellers-etal-2018-swag, zellers-etal-2019-hellaswag}. Following the formulation of the SWAG dataset~\citep{zellers-etal-2018-swag}, we approach next event prediction task as a multiple-choice task, in which a model needs to predict the most likely continuation to a given setting out of the candidate events. For instance, when the prompt goal is ``Sabit Disk Nasıl Çıkarılır? (How to Remove a Hard Drive)'', the prompt step is ``Bilgisayarın kasasını aç. (Open the Computer Case.)'' and the candidate steps are: \\
            A. Bilgisayar kasasının içinde sabit diski bul. (Locate the hard drive inside the computer.) \\
            B. Bilgisayarının verilerini yedekle. (Back up your computer's data.) \\
            C. Masaüstü anakartınla uyumlu bir sabit disk satın al. (Buy a hard drive that is compatible with your desktop motherboard.) \\
            D. Windows yüklü bir masaüstü bilgisayarının olduğundan emin ol. (Make sure that you have a Windows desktop computer.) \\
            then the answer would be \textbf{A}. \\
    With the subgroup of our corpus labeled as ordered, we iteratively collect the prompt goals and two consecutive steps to use the prior step as the prompt step and the later step as the positive candidate. After obtaining the positive candidate, we use a similar sampling strategy that we used for goal inference. Unlike in goal inference, we additionally take pronoun token embeddings into account in order not to break the coreference chains.
    
\begin{figure}[h]
            \centering
            \includegraphics[width=0.5\textwidth]{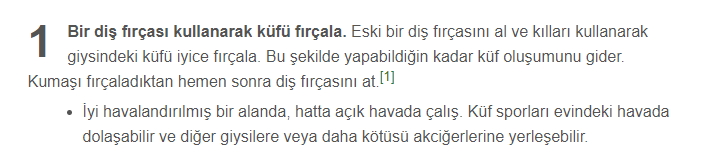}
            \caption{An example step from the ``How to Get Mold Out of Clothing'' tutorial. The bolded part is the step headline, used as the summary, while the step description serves as the text to be summarized. The step description does not include the step headline, formulating the summarization task as the abstractive summarizaton.}
            \label{figure:step_headline_and_description}
            \end{figure}

    \paragraph{Summarization} Similar to \citet{ladhak-etal-2020-wikilingua, Koupaee2018WikiHowAL}, we formulate it as an abstractive summarization. We follow the data format proposed by \citet{Koupaee2018WikiHowAL} and build on the WikiLingua's \citep{ladhak-etal-2020-wikilingua} contributions to performing summarization over Turkish procedural text. Within the wikiHow platform, every step is composed of a concise headline resembling a summary and a descriptive paragraph providing detailed information about the step. In cases where tutorials lack methods or parts, we use the descriptions and headlines of the steps to form two distinct text bodies. These text bodies are then utilized to generate document-summary pairs. In the tutorials containing methods or parts, we follow a similar approach at the method or part level. An illustration of a step from the tutorial "Giysiden Küf Nasıl Çıkarılır? (How to Get Mold Out of Clothing)" is presented in Figure~\ref{figure:step_headline_and_description}.

    \subsection{Test Split Construction via Expert Annotation}

    Despite being synthetic, we incorporate examples from the machine-translated portion of our corpus into the test splits of our datasets. This decision stems from the limited availability of intersecting how-to tutorials on similar topics within the original Turkish wikiHow. Consequently, sampling negative candidates with high semantic similarity becomes challenging, leading to easily distinguishable positive candidates.


    Due to the automated nature of our dataset creation process, some noise is present in the multiple choice task datasets. This noise includes false negative candidates and translations that are incorrect or ambiguous. For instance, consider the step ``Yarayı tedavi etmeden önce ve sonra uygun el yıkama yapın. (Perform proper hand washing before and after treating the wound.)'' which has a positive candidate of ``Drenaj Yarasını Tedavi Etmek (Treat a Draining Wound)'' and a negative candidate of ``Yatak Yaralarını Tedavi Etmek (Treat Bedsores).'' While the negative candidate is sampled due to its high semantic similarity with the positive candidate, it is also a plausible option for the given step. To address this issue, we employ expert annotation to validate the test splits of the multiple choice datasets and eliminate such noisy examples.

    We randomly sample 1000 examples for each of goal inference, step inference, and next event prediction tasks and 1500 examples for step ordering tasks, to be annotated by two experts. Firstly, the experts verify if there are multiple plausible candidates for each example. Secondly, the experts examine whether the translation has altered the meaning of any candidate. The annotation process results in approximately 60-80\% of the randomly sampled examples, which are later utilized as the test splits, as illustrated in Table~\ref{table:task_specific_datasets}.

 

            
\section{Models}

Due to the distinct formulation of each task, we describe them individually below. For each task, we define the overall methodology. The implementation settings are described in Appendix \ref{sec:imp_details}.

\subsection{Linking Actions}
        We employ the retrieve-then-rerank strategy proposed by \citet{zhou-etal-2022-show}. As the name suggests, retrieve-then-rerank approach consists of two stages: i) Retrieval: the steps and goals are encoded in the dense embeddings space to perform semantic similarity search, and ii) Reranking: the top-n candidate goals are reranked for a given step by jointly encoding them.


        During the retrieval stage, we initially encode the steps and goals individually. By obtaining embeddings of the steps and goals, we proceed to calculate the cosine similarity between pairs of goals and steps. Leveraging these computed cosine similarities, we employ semantic similarity search with FAISS~\citep{johnson2019billion} to retrieve the top-n most similar candidates for each step. We experiment with both dense and sparse retrieval (e.g., BM25 \citep{bm25}). For dense retrieval, we experiment with various sentence embedding models with different architectures (e.g., bi-encoder, cross-encoder), different fine-tuning data (e.g., NLI, STS, or both), and different pretraining data (e.g., Turkish or multilingual) described in details at Appendix~\ref{ssec:appendix_link}. In addition to existing sentence embeddings, we inspire by the recent success of the SimCSE architecture~\citep{gao-etal-2021-simcse}, and train our own Turkish-specific sentence embedding model, SimCSE-TR, in several training stages utilizing the text from Turkish Wikipedia and Turkish NLI~(see Appendix~\ref{sec:training_simcse_tr}). Since each step has only one ground-truth goal, we use the standard recall metric to evaluate the retrieval models.  

        Encoding steps and goals independently is efficient; however, might result in information loss. Therefore, we rerank the top-n candidate list for each step, considering the step itself, the candidate goal, and the step's context, which includes surrounding steps or its goal. To accomplish this, we concatenate and input them into another model, utilizing the \textsc{[CLS]} token in the final hidden state to calculate a second similarity score. By reordering the top-n candidates based on the second similarity scores, we obtain the final list. 
        
    \subsection{Multiple Choice Tasks}
        Since the goal inference, step inference, step ordering, and next event prediction tasks share a consistent formulation and adhere to the data format of the SWAG~\citep{zellers-etal-2018-swag} dataset, we employ an identical methodology across these tasks.
    

        
        The models we investigate utilize a common strategy for the aforementioned multiple choice tasks. We provide the models with a question---the goal text for step inference and step ordering, the step text for goal inference, and both for next event prediction. Alongside the question, the models are given a candidate answer from the multiple options and generate a logit for that particular candidate. During the training process, we employ the cross-entropy loss to fine-tune our models, aiming to predict the correct candidate. We experiment with both Turkish-specific (i.e. BERTurk and DistilBERTurk \citep{stefan_schweter_2020_3770924}) and multilingual (i.e. XLM \citep{conneau-etal-2020-unsupervised}) Transformer encoder models, as described in Appendix~\ref{sec:appendix_multi}. We use the standard metric, accuracy, to measure the performance. In addition to fine-tuning, we employ the models in a zero-shot setting. 
        
    \subsection{Summarization}
         \citet{safaya-etal-2022-mukayese} introduces large pre-trained text generation models fine-tuned on the Turkish news summarization datasets, presenting out-of-domain baselines for summarization. We further fine-tune the aforementioned models to generate the short descriptions (summaries) of the procedural tutorials (longer text bodies). We then test both the out-of-domain and in-domain procedural summarization models. Similarly, we experiment with both language-specific decoder models such as TR-BART \citep{safaya-etal-2022-mukayese}, and multilingual decoder models such as mBART \citep{liu-etal-2020-multilingual-denoising} and mT5 \citep{xue-etal-2021-mt5}, described in Appendix \ref{appendix:summar}. We use the standard ROUGE metrics for evaluation.

\section{Results and Discussion}
\label{section:results}

    \subsection{Linking Actions}
        
        We give the main results for both the retrieval and reranking models in Table \ref{table:linking_actions_results}. We observe that our SimCSE-TR models discussed in Appendix \ref{sec:training_simcse_tr} outperform other baselines by a large margin. Furthermore, multilingual models generally perform worse than Turkish-specific models, which is expected. Similarly, XLM-R based models trained on parallel data for 50 languages \citep{conneau-etal-2020-unsupervised} generally perform worse than BERTurk-based models. Finally, we find that BM25 cannot be used in practical scenarios due to its low performance. 


        In the reranking stage, we introduce the ground-truth goal into the candidates' list, initially generated by the top-performing retrieval model. This addition occurs randomly after the 10th candidate, allowing us to assess the impact of reranking models. This modification significantly enhances the R@10 metric. However, it is noteworthy that DistilBERTurk exhibits a decline in R@1 performance, indicating that while it can distinguish the ground truth goals from other candidates, its improvement is limited to R@10. Conversely, BERTurk demonstrates a boost in both R@1 and R@10 performances.


        The top-performing Turkish retrieval model achieves a comparable performance to the best-performing English retrieval model examined in \citet{zhou-etal-2022-show}. We attribute this similarity to the fact that the effectiveness of semantic similarity search remains consistent when the data and model quality levels are comparable across languages. However, it is worth noting that the best-performing Turkish reranking model exhibits a noticeable decline in performance compared to its English counterpart. We speculate that two factors contribute to this discrepancy: firstly, English dataset is significantly larger than Turkish dataset (21K vs. 1.7K), and secondly, the best-performing English reranking model, DeBERTa \citep{deberta}, is larger in size compared to the best-performing Turkish reranking model, BERTurk.

        \begin{table}[]
        \resizebox{\columnwidth}{!}{%
        \begin{tabular}{lrrr}
        \toprule
        \textbf{Model} & \multicolumn{1}{l}{\textbf{R@1}} & \multicolumn{1}{l}{\textbf{R@10}} & \multicolumn{1}{l}{\textbf{R@30}} \\
        \midrule
        XLM-R+NLI+STS & 0.2 & 0.9 & 1.1 \\
        BM25 & 4.5 & 13.4 & 18.4 \\
        BERTurk+NLI+STS & 9.3 & 17.3 & 24.3 \\
        Unsup. SimCSE-TR\small{XLM-R} & 11.6 & 24.5 & 33.9 \\
        XLM-R-XL-Paraphrase & 15.9 & 33.0 & 41.1 \\
        S-XLM-R+NLI+STS & 17.0 & 31.6 & 40.7 \\
        LaBSE & 19.8 & 32.0 & 40.0 \\
        Sup. SimCSE-TR\small{XLM-R} & 25.9 & 42.7 & 54.1 \\
        S-BERTurk+NLI+STS & 27.3 & 47.7 & 55.7 \\
        Unsup. SimCSE-TR\small{BERTurk} & 31.4 & 52.0 & 61.4 \\
        \textbf{Sup. SimCSE-TR\small{BERTurk}} & \textbf{33.4} & \textbf{55.7} & \textbf{67.3} \\
        \bottomrule
        \quad + DistilBERTurk & 30.7 & 74.8 & \textemdash \\
        \quad + \textbf{BERTurk} & \textbf{40.5} & \textbf{78.9} & \textemdash \\
        \bottomrule
        \end{tabular}%
        }
        \caption{The R@n indicates the percentage of the ground-truth goal being in the top-n candidates for a given step. The last two rows show the performances of the reranker models after including the gold goals in top-30 candidates generated by the best performing model, while the rest is retrieval only. We discuss the baseline models in Appendix \ref{sec:baselines}.}
        \label{table:linking_actions_results}
        \end{table}
        
    \subsection{Multiple Choice Tasks}
        We observe a common pattern for the goal inference, step inference, and next event prediction tasks\footnote{While we manually check the performances of models with different random seeds, we only report the best run for all models, since the observed variances among different runs are small and would not cause any change in the rankings.}: BERTurk performs the best, XLM-R is a close runner-up to the BERTurk, and DistilBERTurk performs slightly worse than XLM-R, as given in Table \ref{table:mc_tasks}. In step ordering, DistilBERTurk performs slightly better than XLM-R.

        Zero-shot performances of these models are on par with the random chance of guessing correctly, which means they cannot inherently understand the relationships between goal and step pairs, as well as step and step pairs. Furthermore, zero-shot performances of XLM-R are noticeably worse than those of BERTurk and DistilBERTurk. We believe this is due to the multilingual nature of XLM-R, which is not specialized in Turkish, unlike BERTurk and DistilBERTurk.
        

        Significant improvements are observed with the fine-tuned models. The fine-tuned XLM-R model outperforms the fine-tuned DistilBERTurk model in all multiple choice tasks, except for step ordering. This observation suggests that the XLM-R model not only enhances its ability to select the correct candidate but also improves its understanding of the Turkish language through fine-tuning.

        

        When comparing the performance of language-specific models trained on Turkish data to those trained on English data, noticeable differences are observed. Turkish models exhibit significantly lower performances in goal inference and step ordering tasks. We attribute these variations to the dissimilarity in our sampling strategy, as explained in \S\ref{subsection:downstream}. Our sampling strategy considers a broader range of parts of speech compared to the approach used by \citet{zhang-etal-2020-reasoning}, resulting in candidates that are more similar at the embedding level and thereby increasing the difficulty. Additionally, while the performance decreases in goal inference, there is a slight improvement in step inference. This can be attributed to the fact that goals typically consist of less diverse parts of speech, mostly composed of a noun and a verb. As a result, the candidates sampled for goal inference tend to be more similar at the embedding level compared to step inference candidates, which often include additional parts of speech such as adjectives and adverbs.

        
        Although we do not practice adversarial filtering to create our next event prediction dataset, we believe our sampling strategy also presents its own challenges. While the results shared in \citet{zellers-etal-2018-swag, zellers-etal-2019-hellaswag} are significantly lower than those of our models, the leaderboards for SWAG\footnote{\href{https://leaderboard.allenai.org/swag/submissions/public}{https://leaderboard.allenai.org/swag/submissions/public}} and HellaSwag\footnote{\href{https://rowanzellers.com/hellaswag/}{https://rowanzellers.com/hellaswag/}} datasets show that the challenge adversarial filtering brings can be overcome. Considering these, our results given in Table \ref{table:mc_tasks} are significantly lower than their English counterparts, suggesting a large room for improvement.

        Additionally, we evaluate out-of-domain performances of some best-performing models to better understand their abilities in procedural tasks and find out their performances are generalizable to a certain extent, as discussed in Appendix \ref{sec:cross_test}.
        
        \begin{table}
            \begin{center}
            \scalebox{0.65}{
            \begin{tabular}{lrrrr}
                \hline
                \multirow{2}{*}{Task} &
                {Goal} &
                {Step} &
                {Step} &
                {Next Event} \\
                & Inference & Inference & Ordering & Prediction \\
                \toprule
                Random & 25.00 & 25.00 & 50.00 & 25.00 \\
                \midrule
                XLM-R ZS (125M) & 22.70 & 23.86 & 42.90 & 25.65 \\
                DistilBERTurk ZS (66M) & 25.81 & 24.51 & 47.01 & 27.02\\
                \textbf{BERTurk ZS} (110M) & \textbf{26.52} & \textbf{27.45} & \textbf{49.46} & \textbf{32.82}\\
                \midrule
                DistilBERTurk FT (66M) & 66.19 & 85.78 & 70.13 & 83.66 \\
                XLM-R FT (125M) & 69.30 & 87.42 & 68.17 & 85.95 \\
                \textbf{BERTurk FT} (110M) & \textbf{72.40} & \textbf{91.34} & \textbf{72.09} & \textbf{88.55} \\
                \bottomrule
              \end{tabular}}
            \caption{Zero-Shot and Fine-Tuned performances of XLM-R, DistilBERTurk, and BERTurk models on multiple choice tasks, evaluated using accuracy. FT indicates that the model is fine-tuned on the task-specific data and ZS indicates zero-shot performance.}
            \label{table:mc_tasks}
            \end{center}
        \end{table}

    \subsection{Summarization}

        The results are given in Table \ref{table:summ_results}. As anticipated, in the summarization task, models that are fine-tuned on procedural summarization data outperform their out-of-domain fine-tuned counterparts. However, the performance improvement observed is relatively modest. We attribute this to the fact that the out-of-domain models still possess a robust capability acquired through their prior training on news summarization tasks.
        

        Additionally, the multilingual out-of-domain models demonstrate superior performance compared to the single Turkish-specific model, TR-BART. However, in the procedural summarization task, TR-BART exhibits a higher performance boost and performs marginally better than procedural mT5. Both out-of-domain and procedural mBART models outperform other models. We attribute this to substantial size difference of mBART, which gives it an advantage over the other models.
        
        
        When taking into account the model sizes and their multilingual capabilities, we conclude that both the specialization to Turkish and larger model sizes contribute to the overall performance improvement. However, our analysis reveals that a substantial difference in size can compensate for the multilingual aspect. This is evident in the comparison between out-of-domain and procedural TR-BART and mBART models, as presented in Table \ref{table:summ_results}.
        
        \begin{table}
            \begin{center}
            \scalebox{0.7}{
            \begin{tabular}{lrrr}
            \toprule
            \textbf{Model} & \textbf{ROUGE-1} &\textbf{ROUGE-2} & \textbf{ROUGE-L}\\
            \midrule
            {TR-BART OOD (120M)} & 16.28 & 4.21 & 12.35 \\
            {mT5-base OOD (220M)} & 17.09 & 4.53 & 13.05\\
            {mBART OOD (680M)} & 18.30 & 5.12 & 13.82\\
            \midrule
            {TR-BART PRO (120M)} & 19.59 & 5.64 & 13.68 \\
            {mT5-base PRO (220M)} & 19.30 & 5.33 & 14.42 \\
            \textbf{mBART PRO (680M)} & \textbf{22.62} & \textbf{6.43} & \textbf{15.69} \\
            \bottomrule
            \end{tabular}
            }
            \caption{ Out-of-Domain Fine-Tuned, and Procedural Fine-Tuned performances of TR-BART, mBART, and mT5-base models in summarization task.}
            \label{table:summ_results}
            \end{center}
        \end{table}  
        
\section{Conclusion}
     PLU tasks encompass various skills such as semantic analysis, commonsense reasoning, and coreference resolution. However, PLU has been primarily explored in English and the scarcity of language-specific resources limits its study in other languages. To address this gap, we present a case study in Turkish and introduce a centralized benchmark comprising six downstream tasks on procedural documents. We leverage machine translation tools and implement stringent quality control measures. We curate high-quality task data through language-specific filtering and manual annotation. Our experiments reveal that language-specific models tend to outperform multilingual models, but the model size is a critical factor. Tasks that involve rigorous language-specific preprocessing, such as goal inference, prove to be more challenging. Despite advancements, our best-performing models still lag behind their English counterparts, indicating large room for improvement. We release all resources publicly for further research.
    
    
    
    

\section*{Limitations}
    Our corpus creation method heavily relies on the success of the machine translation systems. However, such systems might have downfalls in specific cases. Local contexts and metrics are examples of such downfalls. We observe that some tutorials from the original Turkish wikiHow are localized, not directly translated. For instance, the Turkish counterpart of the tutorial titled "How to Lose 10 Pounds in 10 Days" is "10 Günde Nasıl 5 Kilo Verilir?" (How to Lose 5 Kilograms in 10 Days). In our case, Google Translate cannot distinguish these nuances. 

    Since the translated portion of our corpus makes up the majority, our models might pick up the translation artifacts, which, in turn, diminishes their success in actually learning their objective tasks. 

    mBART and mT5 models might generate biased summarizations, since they are previously trained on multilingual data and then fine-tuned on news summarizations before being fine-tuned on procedural documents. 

    The heavyweight fine-tuning and inference of mBART and mT5 sets a natural limitation to their usage. However, we overcome this limitation by practicing lightweight alternative solutions, such as half precision floating point format (FP16) training, optimization libraries, and gradient accumulation and checkpointing\footnote{To the best of our knowledge, mT5 models currently cannot be trained with gradient checkpointing.}.

    Lastly, the method we propose for the creation of procedural corpora in low-resource languages is implicitly dependent on the amount of resources for a language. This is because machine translation systems might not work in some low-resource languages as well as they work for Turkish.


\section*{Ethics Statement}
    We use the content of wikiHow, which allows for the usage of its content under limited specific circumstances within the Creative Commons license. We abide all the conditions required by the Creative Commons license. The requirements of the Creative Commons also make the usage of English wikiHow corpus that we translate possible.

    Since the source of the majority of our corpus and datasets are from translated tutorials, they might contain implicit biases due to the translation. Consequently, models trained on such data are also vulnerable to these biases.


\section*{Acknowledgements}
    This work has been supported by the Scientific and Technological Research Council of Türkiye~(TÜBİTAK) as part of the project ``Automatic Learning of Procedural Language from Natural Language Instructions for Intelligent Assistance'' with the number 121C132. We also gratefully acknowledge KUIS AI Lab for providing computational support. We thank our anonymous reviewers and the members of GGLab who helped us improve this paper. We especially thank Shadi Sameh Hamdan for his contributions to setting up the implementation environment.

\bibliographystyle{acl_natbib}
\bibliography{custom}

\appendix
\section{Baselines}
\label{sec:baselines}

\subsection{Linking Actions}
\label{ssec:appendix_link}

\paragraph{S-BERTurk + NLI + STS} is the bi-encoder model that employs the Siamese and ternary network structures \citep{reimers-gurevych-2019-sentence} to  derive close fixed-size sentence embeddings in vector space \citep{beken-fikri-etal-2021-semantic}.
\paragraph{S-XLM-R + NLI + STS} is the bi-encoder model that employs the Siamese and ternary network structures \citep{reimers-gurevych-2019-sentence} to  derive close fixed-size sentence embeddings in vector space \citep{beken-fikri-etal-2021-semantic}.
\paragraph{BERTurk + NLI + STS} is the cross-encoder model that averages the BERT embeddings \citep{beken-fikri-etal-2021-semantic}.
\paragraph{XLM-R + NLI + STS} is the cross-encoder model that averages the XLM-R embeddings \citep{beken-fikri-etal-2021-semantic}.
\paragraph{LaBSE} stands for Language-agnostic BERT Sentence Embedding. It is trained on multilingual data for translation language modeling and produces sentence embeddings for 109 languages, including Turkish \citep{feng-etal-2022-language}. We use the pretrained LaBSE model to generate Turkish sentence embeddings\footnote{\href{https://huggingface.co/sentence-transformers/LaBSE}{https://huggingface.co/sentence-transformers/LaBSE}}.
\paragraph{XLM-RoBERTA-base-XL-Paraphrase} is a XLM-R model \citep{conneau-etal-2020-unsupervised} trained to imitate SBERT-paraphrases on parallel data for 50 languages (including Turkish) using multi-lingual knowledge distillation \citep{reimers-gurevych-2020-making}. We use the pretrained XLM-RoBERTA-base-XL-Paraphrase model to generate Turkish sentence embeddings\footnote{\href{https://huggingface.co/sentence-transformers/paraphrase-xlm-r-multilingual-v1}{https://huggingface.co/sentence-transformers/paraphrase-xlm-r-multilingual-v1}}.
\paragraph{BM25} is a ranking function used to estimate the relevance between a set of documents to a given query based on the query terms appearing in each document \citep{bm25}. We use the BM25+ algorithm from the Rank-BM25 library\footnote{\href{https://github.com/dorianbrown/rank_bm25}{https://github.com/dorianbrown/rank\_bm25}}, which implements the BM25 algorithms from \citep{Trotman2014ImprovementsTB}.

\subsection{Multiple Choice Tasks}
\label{sec:appendix_multi}

\paragraph{DistilBERTurk} is the distilled version of its teacher model BERTurk, trained following the knowledge distillation method introduced by \citet{Sanh2019DistilBERTAD} \citep{stefan_schweter_2020_3770924}.
\paragraph{XLM-R} is a transformer-based model trained on large multilingual data using the objective of multilingual masked language modeling \citep{conneau-etal-2020-unsupervised}.
\paragraph{BERTurk} is a transformer-based model trained on a combination of Turkish web corpora following the training methodology of \citet{devlin-etal-2019-bert} \citep{stefan_schweter_2020_3770924}.

\subsection{Summarization}
\label{appendix:summar}
\paragraph{TR-BART OOD} is a Seq2Seq Transformer \citep{NIPS2017_3f5ee243} trained on the Turkish split of the MLSUM dataset \citep{scialom2020mlsum} following the configuration of BART Base \citep{lewis-etal-2020-bart}\footnote{\href{https://huggingface.co/mukayese/transformer-turkish-summarization}{https://huggingface.co/mukayese/transformer-turkish-summarization}}.
\paragraph{mBART OOD} is a fine-tuned version of the pre-trained mBART50 \citep{liu-etal-2020-multilingual-denoising}. mBART50 is pre-trained on data from 50 different languages, and mBART OOD is fine-tuned on the Turkish split of MLSUM \citep{scialom2020mlsum}\footnote{\href{https://huggingface.co/mukayese/mbart-large-turkish-summarization}{https://huggingface.co/mukayese/mbart-large-turkish-summarization}}.
\paragraph{mT5-base OOD} is a fine-tuned version of the pre-trained mT5-base \citep{xue-etal-2021-mt5}. mT5-base is a multilingual variant of T5 \citep{2020t5} that was pre-trained on a new Common Crawl-based dataset covering 101 languages, and mT5-base OOD is fine-tuned on the Turkish split of MLSUM \citep{scialom2020mlsum}\footnote{\href{https://huggingface.co/mukayese/mt5-base-turkish-summarization}{https://huggingface.co/mukayese/mt5-base-turkish-summarization}}.

\section{Classifying Orderliness of the Tutorials}
\label{sec:classifying_orderliness_of_the_tutorials}
    wikiHow mostly contains two type of tutorials: i) tutorials with consecutive steps that must be followed in sequence (i.e. HDMI Televizyona Nasıl Bağlanır? (How to Connect HDMI to TV) has the steps Televizyonunda kullanılabilir bir HDMI girişi bul. (Locate an available HDMI port on your TV.), Doğru HDMI kablosunu al. (Get the right HDMI cable.), Kablonun bir ucunu cihaza bağla. (Connect one end of the cable to the device.)), ii) tutorials with steps that are parallel or alternative procedures to each other (i.e. Evde Ateş Nasıl Düşürülür? (How to Cure Fever at Home) tutorial has the steps Bol su iç. (Drink lots of water.), Rahat giysiler giy. (Wear comfy clothes.), and Oda sıcaklığını düşür. (Lower the room temperature.)).

    Since the step ordering and next event prediction tasks require tutorials with ordered steps, we need to predict the orderliness of the tutorials in our corpus. First, expert authors annotate 900 tutorials based on the criteria of orderliness. With the obtained data, we fine-tune a BERTurk \citep{stefan_schweter_2020_3770924} model for the binary text classification objective. Finally, we use it to classify each tutorial in our corpus, and use the tutorials labeled as ordered for the step ordering and next event prediction tasks. Our fine-tuned model's performance on our test split can be seen in Table \ref{table:classifying_orderliness_table}.

    \begin{table}[!htp]
        \centering
        \begin{tabular}{rrrr}
            \textbf{Accuracy} & \textbf{Precision} & \textbf{Recall} & \textbf{F1}\\
            \midrule
            86.67 & 85.34 & 90.14 & 87.67  
        \end{tabular}
        \caption{Finetuned BERTurk's performance on our test split. We split the annotated 900 tutorials with the ratio of 70:15:15 (training:evaluation:test).}
        \label{table:classifying_orderliness_table}
    \end{table}

\section{SimCSE-TR}
\label{sec:training_simcse_tr}
    From using them to filter the goal and step inference tasks data to utilizing them in the retrieval stage of the linking actions task, we take advantage of sentence embeddings in a broad range. Therefore, we train a new Turkish-specific sentence embedding model utilizing the SimCSE architecture \citep{gao-etal-2021-simcse}, which we name as SimCSE-TR. 

    SimCSE architecture employs a contrastive learning objective to derive meaningful sentence embeddings, with the hidden dropout mask acting as a minimal data augmentation method. In the unsupervised setting, SimCSE uses sentences from English Wikipedia to sample positive pairs by generating the representations of the same sentence with different dropout masks and negative pairs with the representations of different sentences. In the supervised setting, it integrates the annotated sentence pairs from natural language inference datasets into its contrastive training objective, utilizing the ``entailment'' pairs as positive pairs and ``contradiction'' pairs as hard negative pairs \citep{gao-etal-2021-simcse}. Compared to other sentence embedding models and architectures, SimCSE converges faster with fewer data, which makes it lightweight to train and use. Furthermore, with a better aligned and more uniform latent space, it performs better on semantic textual similarity tasks and generates more distinguishable representation for sentences. 
    
    Following the implementation in SimCSE, we use randomly sampled one million sentences from Turkish Wikipedia for the unsupervised setting and the Turkish NLI datasets \citep{budur-etal-2020-data} for the supervised setting to train BERTurk \citep{stefan_schweter_2020_3770924} and XLM-R based Turkish SimCSE models \citep{conneau-etal-2020-unsupervised}. Similar to the English SimCSE, we train the unsupervised models for 1 epoch and the supervised models for 3 epochs. For each of the settings, we carry out a grid-search of batch size$\in$\{64, 128, 256, 512\}, learning rate$\in$\{$1e-5$, $3e-5$, $5e-5$\}, and maximum sequence length$\in$\{16, 32, 64\} on Turkish STS-B development set, and report the best combinations in Table \ref{table:simcse_implementation_details}. We use the edited version of the SentEval \citep{conneau-kiela-2018-senteval} library shared in SimCSE Github repository\footnote{\href{https://github.com/princeton-nlp/SimCSE}{https://github.com/princeton-nlp/SimCSE}} for the testing, and share the results in Table \ref{table:simcse_test_metrics}. Although they are not trained or fine-tuned on the train split of the Turkish STS-B, SimCSE-TR models perform comparably to other Turkish-specific sentence embedding models that are trained on Turkish STS-B. 
    \begin{table}
        \centering
        \scalebox{0.7}{
            \begin{tabular}{lrrrr}
                \hline
                \multirow{2}{*}{Hyperparameter} &
                \multicolumn{2}{c}{Unsupervised} &
                \multicolumn{2}{c}{Supervised} \\
                & BERTurk & XLM-R & BERTurk & XLM-R \\
                \hline
                Batch Size & 64 & 512 & 512 & 512 \\
                Learning Rate & 1e-5 & 1e-5 & 3e-5 & 5e-5 \\
                Max. Seq. Length & 64 & 64 & 64 & 64 \\
                \hline
            \end{tabular}
        }
        \caption{Hyperparameters used in the training of SimCSE-TR models.}
        \label{table:simcse_implementation_details}
    \end{table}

    \begin{table}
            \begin{center}
            \scalebox{0.8}{
            \begin{tabular}{lrrr}
            \toprule
            & \textbf{Pearson} & \textbf{Spearman}\\
            \midrule
            Unsup SimCSE-TR XLM-R & 66.23 & 66.95 \\
            Unsup. SimCSE-TR BERTurk & 74.31 & 72.56\\
            S-XLM-R\small{$\heartsuit$} + NLI + STS & 77.26 & 77.32 \\
            Sup SimCSE-TR XLM-R & 79.70 & 80.30 \\
            Sup. SimCSE-TR BERTurk & 79.07 & 81.06 \\
            XLM-R\small{$\heartsuit$} + NLI + STS & 81.94 & 81.21 \\
            S-BERTurk\small{$\heartsuit$} + NLI + STS & 82.85 & 83.31 \\
            \textbf{BERTurk\small{$\heartsuit$} + NLI + STS} & \textbf{85.36} & \textbf{84.59} \\
            \bottomrule
            \end{tabular}
            }
            \caption{Performances of SimCSE-TR and other Turkish-specific sentence embedding models on the test split of the Turkish STS-B. $\heartsuit$: taken directly from \citep{beken-fikri-etal-2021-semantic}. Pearson and Spearman correlations were reported as $\rho \times 100$.}
            \label{table:simcse_test_metrics}
            \end{center}
    \end{table}

\section{\textsc{ÇeVeri}}
\label{sec:ceveri}
    \textsc{ÇeVeri} utilizes the pandas library \citep{McKinney2011pandasAF} and recursive search to detect text values in seven different file format, .txt, .json, .xlsx, .csv, .xml, .pkl, and .docx. It, then, uses Google Translate to translate and replace detected texts. Although there is no usage-limit set by \textsc{ÇeVeri}, employment of the Google Translate makes it optimal to use \textsc{ÇeVeri} for a dataset consisting of a high number of smaller files, instead of a dataset consisting of a lower number of bigger files. Since it uses Google Translate in its backend, \textsc{ÇeVeri} can translate not only English but also all the languages Google Translate supports, as well as detecting and translating from unknown source languages.

\section{Investigating the Feasibility of the Usage of Machine-Translated Data}
\label{sec:feasibility_mt}
    In order to analyze the feasibility of using machine-translated data for studying procedural tasks, we conduct a pilot study in linking actions task. 

    First, we shuffle and recreate the train and test splits of our linking actions dataset. However, we do not include any silver data in the test split this time, contrary to what we did in \S\ref{subsection:downstream}. To test the practicability of using silver data, we incrementally increase the amount of the machine-translated data in the train split. We train the reranking models on these train splits with varying amounts of silver data and test them on the test split that solely consists of gold data. As seen in Figure \ref{fig:ablation_study_on_translated_data}, the utilization of silver data brings a noticeable improvement over the usage of only gold data to the performance of the reranking model. Furthermore, reranking models trained with a combination of gold and silver data outperforms the retrieval model consistently, on the contrary of reranking model trained with solely gold data underperforming the retrieval model in R@1 performance.

    \pgfplotsset{width=7cm,compat=1.9}
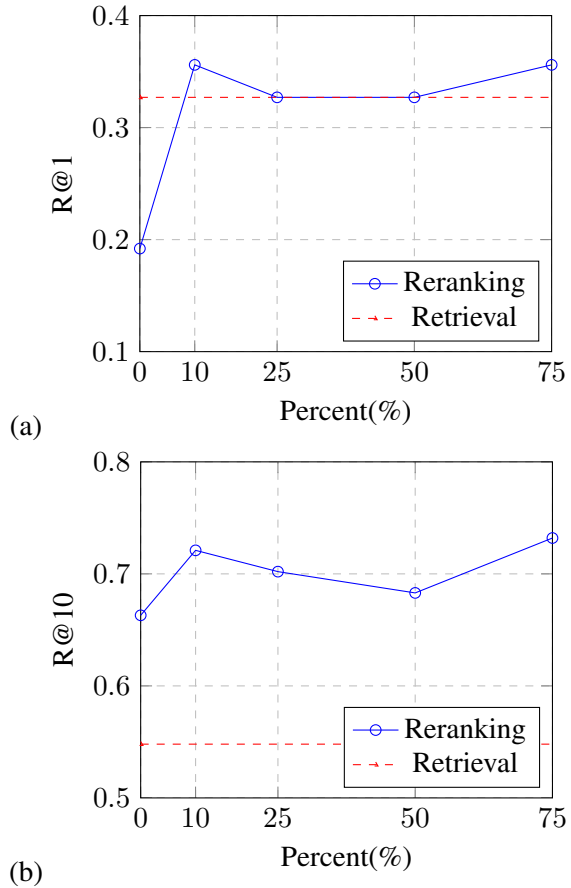
\begin{figure}[ht]
\begin{centering}
(a)\begin{tikzpicture}
\begin{axis}[
    title={},
    xlabel={Percent(\%)},
    ylabel={R@1},
    xmin=0, xmax=75,
    ymin=0.1, ymax=0.4,
    xtick={0,10,25,50,75},
    ytick={0.1, 0.2, 0.3, 0.4},
    legend pos=south east,
    ymajorgrids=true,
    xmajorgrids=true,
    grid style=dashed,
]
\addplot[
    color=blue,
    mark=o,
    mark size=2pt,
    ]
    coordinates {
    (0,	    0.192)
    (10,	0.356)
    (25,	0.327)
    (50,	0.327)
    (75,	0.356)
    };
    \addlegendentry{Reranking}
\addplot[
    color=red,
    mark=triangle,
    mark size=1pt,
    dashed,
    ]
    coordinates {
    (0,	    0.327)
    (100,	0.327)
    };
    \addlegendentry{Retrieval}
\end{axis}
\end{tikzpicture}
(b)\begin{tikzpicture}
\begin{axis}[
    title={},
    xlabel={Percent(\%)},
    ylabel={R@10},
    xmin=0, xmax=75,
    ymin=0.5, ymax=0.8,
    xtick={0,10,25,50,75},
    ytick={0.5, 0.6, 0.7, 0.8},
    legend pos=south east,
    ymajorgrids=true,
    xmajorgrids=true,
    grid style=dashed,
]
\addplot[
    color=blue,
    mark=o,
    mark size=2pt,
    ]
    coordinates {
    (0,	    0.663)
    (10,	0.721)
    (25,	0.702)
    (50,	0.683)
    (75,	0.732)
    };
    \addlegendentry{Reranking}
\addplot[
    color=red,
    mark=triangle,
    mark size=1pt,
    dashed,
    ]
    coordinates {
    (0,	    0.548)
    (100,	0.548)
    };
    \addlegendentry{Retrieval}
\end{axis}
\end{tikzpicture}
\caption{Performances of the BERTurk-based reranking models trained with different percentages of the translated data's train split. a) shows the performance change on R@1 and b) on R@10. 0\% means reranking model is trained only with the originally Turkish data.}
\label{fig:ablation_study_on_translated_data}
\end{centering}
\end{figure}
    
\section{Out-of-Domain Evaluation}
\label{sec:cross_test}
    To better understand the extent of our models' abilities in procedural tasks, we evaluate some of the best-performing models across other tasks.

    Since the system needs to bring a continuation to the given set of actions in the next event prediction task, we hypothesize that next event prediction task implicitly covers the step inference task. In this regard, we believe that next event prediction models learn the relationship between the goals and steps, because the following actions to a given context must simultaneously serve the given goal. To investigate our claim, we test the BERTurk Next Event Prediction model on the test split of our step inference task. As Table \ref{table:cross_test} shows, BERTurk Next Event Prediction model achieves much higher performances than all the zero-shot models and the random probability. 

    To further examine the relationship between the next event prediction and step inference tasks, we also test the BERTurk Step Inference model on the test split of our next event prediction task. As Table \ref{table:cross_test} shows, BERTurk Step Inference model outperforms all zero-shot performances and the random probability, and performs closely to fine-tuned DistilBERTurk NEP, XLM-R NEP, and BERTurk NEP models.

    We believe the lower performance of BERTurk NEP on step inference data than the performance of BERTurk SI on next event prediction data is because BERTurk NEP is fine-tuned in a way that makes it dependent on the context information, which is absent in the step inference task. Similarly, we believe BERTurk SI obtains higher scores on next event prediction data than does BERTurk NEP on step inference data, because next event prediction task provides the context information, which might ease the objective of choosing the positive candidate.

        \begin{table}
            \begin{center}
            \scalebox{1}{
            \begin{tabular}{lrr}
            \toprule
            \textbf{Model} & \textbf{SI} & \textbf{NEP} \\
            \midrule
            {Random} & 25.00 & 25.00 \\
            BERTurk Zero-Shot & 27.45 & 32.82 \\
            BERTurk NEP & 61.93 & 88.55 \\
            BERTurk SI & 91.34 & 80.46 \\
            \bottomrule
            \end{tabular}
            }
            \caption{Performances of the best-performing Step Inference and Next Event Prediction models across step inference and next event prediction tasks. SI: Step Inference, NEP: Next Event Prediction.}
            \label{table:cross_test}
            \end{center}
        \end{table}
    
\section{Implementation Details}
\label{sec:imp_details}
    We implement the reranking models as they are in \citet{zhou-etal-2022-show}'s Github repository\footnote{\href{https://github.com/shuyanzhou/wikihow_hierarchy}{https://github.com/shuyanzhou/wikihow\_hierarchy}} with the same training setting. Since the dataset is small, training of the reranking models are quite lightweight, taking 15 to 45 minutes to train.

    We implement the summarization and multiple choice models using the Hugging Face libraries: Transformers \citep{wolf-etal-2020-transformers}, accelerate \citep{accelerate}, datasets \citep{lhoest-etal-2021-datasets}, and evaluate. Transformers library enables us to work with the pre-trained models, accelerate library eases and accelerates the fine-tuning process and makes it more efficient, datasets library makes it easier to load and use datasets, and evaluate library facilitates the evaluation of the models.

    With the accelerate library, we use FP16 training, gradient accumulation and checkpointing, and the Adafactor loss \citep{Shazeer2018AdafactorAL}. This combination enables fine-tuning all the models on four NVIDIA T4s and test them on only one NVIDIA T4. In this setting, step inference and next event prediction models take 15 to 45 minutes, goal inference models take 30 to 90 minutes, step ordering models take 1 to 3 hours, and summarization models take approximately 9 to 18 hours to fine-tune. 
    
    Since we work with various models across different tasks, the hyperparameter setups for each dedicated task is given in details at \url{\repourl}. 
\end{document}